# DIFFUSION MODELS FOR INTERFEROMETRIC SATELLITE APERTURE RADAR

ALEXANDRE TUEL[*,†], THOMAS KERDREUX [*,†], CLAUDIA HULBERT[†], AND BERTRAND ROUET-LEDUC[†,‡]

ABSTRACT. Probabilistic Diffusion Models (PDMs) have recently emerged as a very promising class of generative models, achieving high performance in natural image generation. However, their performance relative to non-natural images, like radar-based satellite data, remains largely unknown. Generating large amounts of synthetic (and especially labelled) satellite data is crucial to implement deep-learning approaches for the processing and analysis of (interferometric) satellite aperture radar data. Here, we leverage PDMs to generate several radar-based satellite image datasets. We show that PDMs succeed in generating images with complex and realistic structures, but that sampling time remains an issue. Indeed, accelerated sampling strategies, which work well on simple image datasets like MNIST, fail on our radar datasets. We provide a simple and versatile open-source `https://github.com/thomaskerdreux/PDM_SAR_InSAR_generation` to train, sample and evaluate PDMs using any dataset on a single GPU.

## 1. INTRODUCTION

Probabilistic Diffusion Models (PDMs) are a recent family of deep generative models which have demonstrated state-of-the-art performance in image translation [*e.g.*, SWB21] and generation [*e.g.*, DN21, MFNK[+]22] in terms of target distribution coverage and sample quality with respect to other generative models, such as Generative Adversarial Networks (GANs) [GPAM[+]20], Variational Auto-Encoders (VAEs) [KW13, RMW14], autoregressive models [VDOKK16] or Normalizing Flows [DSDB16, PNR[+]21]. In addition, PDMs have the major advantage of being very versatile. They are less prompt to various failures often encountered with other generative approaches, such as *mode collapse* during the training of GANs or *posterior collapse* for VAEs [LTGN19]. Consequently, this considerably reduces the engineering work required to train generative models, and paves the way for fully automated data analysis pipelines in remote sensing.

Intuitively, PDMs consist in destroying data samples by random noise addition, and in learning to reverse this process by denoising images. PDMs [HJA20, ND21] consist of two Markov chains. The *forward* process is a handcrafted Markov chain that iteratively transforms data into noise by injecting simple noise patterns. The *reverse process* is a Markov chain with the same states as the forward process, in which transitions are parametrized by a neural network and trained to reconstruct the data from the noise. Once trained, the reverse process becomes a powerful generative model that has shown state-of-the-art performances on natural images, on natural language processing [AJH[+]21, HNJ[+]21] or in computational chemistry [LSP[+]22, LJH22].

PDMs have only very recently been applied to remote sensing challenges, such as super-resolution [LYP[+]22], change detection [BNP22, GCBGNP22], cloud removal [ZJ23], despeckling [ZZD[+]23] or image fusion [CCW[+]23]. They remain somewhat unknown in the remote sensing community, especially compared to competing models like GANs [JCPA22, LLWH22, WBS22]. In particular, to our knowledge, PDMs have not yet been leveraged for radar remote sensing imagery, *i.e.*, Satellite Aperture Radar (SAR) or Interferometric Satellite Aperture Radar (InSAR). In recent years, deep learning algorithms have been developed to perform a wide range of tasks in radar imagery, such as *e.g.*, SAR image classification, unwrapping of InSAR interferograms [SGG20] or denoising of InSAR time series [RLJD[+]21]. Having access to large training databases is key to the performance of such automated algorithms, and this is where generative models like PDMs may be instrumental. For instance, GANs have been widely applied to augment small, handcrafted SAR datasets and improve the performance of classification and detection of SAR images, [*e.g.*, LZL+18, BSKSGV20, ZZW+20, SXZJ21].

---

[*]Equal contribution (randomized order).
[†] Geolabe LLC, Los Alamos, NM, USA.
[‡] DPRI, Kyoto University, Japan.

The main bottleneck of PDMs is that the original sampling approach for the reverse process [SDWMG15, HJA20] is much slower than for other generative models. Indeed, it consists in iteratively inferring the output from a neural network at each time step of the reverse diffusion process, and is therefore not parallelizable. Since generating images with PDMs may require thousands of diffusion steps on datasets like CIFAR10 or CelebA-HQ to produce high-quality samples [HJA20], the sampling time is orders of magnitude larger than that of competing approaches like GANs.
Nevertheless, since PDMs were first introduced, research has been active to design faster sampling schemes, to strengthen their theoretical understanding, or to improve the training procedure for the reverse processes. All these have already been discussed at length in the literature, see various review papers [*e.g.*, YSM22, Luo22] for more details. In particular, many efficient approximate sampling approaches of the reverse process have been developed that alleviate much of the sampling bottleneck (see Section 2.2 for a more detailed discussion).

Here, we explore the applicability of PDMs to generate realistic radar-based remote sensing imagery, in the context of three key tasks associated with the processing and interpreting of radar data: SAR image classification from backscatter retrievals, unwrapping of InSAR interferograms, and denoising of InSAR-based ground deformation scenes. We begin in Section 2 with an overview of PDMs and associated sampling and evaluation strategies, before presenting the datasets and experiments we conduct in Section 3. We show our results in Section 4 before discussing PDMs more generally in the context of remote sensing data in Section 5.

## 2. METHODS

2.1. **Overview of PDMs.** The goal of generative models is to learn to model the distribution $q(x_0)$ of an ensemble of observed samples $x_0$. The core idea behind PDMs is to iteratively "destroy" an original sample by adding random Gaussian noise to it until nothing but noise remains, and to learn an approximate version of the reverse of this process with a deep learning model. More specifically, PDMs make use of two different Markov chains that share the same states, $(x_t)_{t=0;\cdots;T}$ (but in reverse order), to represent the forward (or noising) process, and the approximate reverse process (Figure 1). The forward process is defined by the following transitions:

$$q(x_t|x_{t-1}) := \mathcal{N}\Big(x_t; \sqrt{1-\beta_t}x_{t-1}, \beta_t I\Big),$$

according to a given *beta schedule* $(\beta_0, ..., \beta_T)$ which determines the characteristics of the noise that is added at each time step of the diffusion process. Variables $(x_t)_{t>0}$ are latent variables of the diffusion process. This Markov chain thus consists in iteratively adding noise to a data sample until the distribution of the last latent variable, $x_T$, is approximately an isotropic Gaussian distribution, given a large enough time step $T$ and a well-chosen beta schedule (see the example at the top row of Figure 2). Being able to reverse that process would therefore provide a simple and convenient approach to sample from the target data distribution (a difficult task a priori) by sampling from a Gaussian distribution (an easy task to perform).

PDMs hence seek to approximate the reverse of this forward process, by learning another Markov chain $(x_t)$ where the moments of the Gaussian Markov transitions $p_\theta(x_{t-1}|x_t)$ are parametrized by a deep learning model, namely:

$$p_\theta(x_{t-1}|x_t) := \mathcal{N}\Big(x_{t-1}; \mu_\theta(x_t, t), \Sigma_\theta(x_t, t)\Big).$$

This second Markov chain has been called a *reverse diffusion process* or *backward process* (see Figure 1 for an example from SAR imagery). New data samples can then be generated by first sampling from a Gaussian distribution $x_T \sim p(x_T)$ and by iteratively applying the parametrized approximate reverse process to gradually remove the noise until $t = 0$. In the original formulation of PDMs [SDWMG15, HJA20], only the first-order moment $\mu_\theta(x_t, t)$ of the Gaussian Markov transition was learned. The covariance matrix $\Sigma_\theta(x_t, t)$ was pre-defined through the beta schedule. Nichol and Dhariwal [ND21] later showed that learning the variance along with the mean allowed to use fewer diffusion time steps and ultimately led to faster sampling.

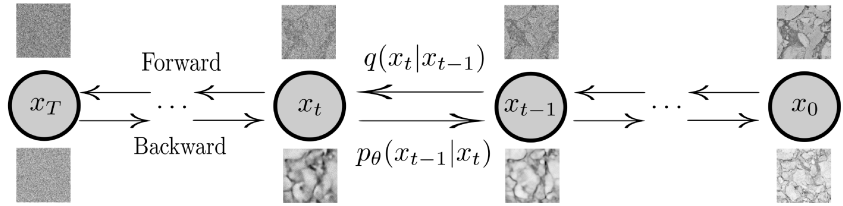

FIGURE 1. Illustration of the forward and learned reverse process. $q(\cdot|\cdot)$ is the handcrafted Markov transition for the *forward* process and $p_\theta(x_{t-1}|x_t)$ is the deep learning parameterized Markov transition of the *reverse/backward* process.

The parameters of the reverse process are learned by minimizing a mean-squared error objective on the predicted noise, or a variational lower bound on the likehood of the latent variables $(x_t)$ according to the parametrized reverse process distribution [HJA20]. Many methods and variations of this approach exist and have been discussed extensively in various survey papers [YSM22, Luo22]. Here, we adopt the hybrid loss (weighted sum between the mean-squared error and variational lower bound objectives) with importance sampling proposed by Nichol and Dhariwal [ND21]. Importance sampling here consists in calculating the loss not at all diffusion time steps, but only on a subset of them according to their respective loss at the previous time step.

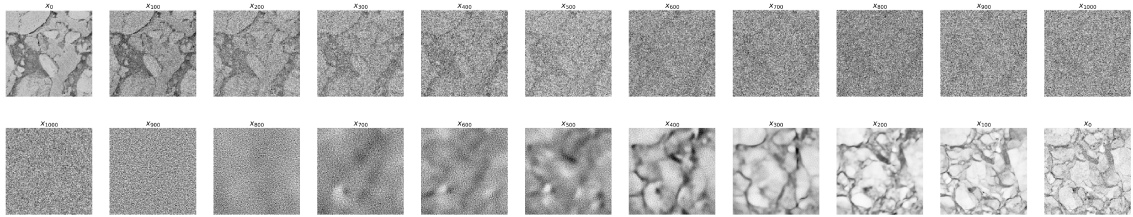

FIGURE 2. An example of noising diffusion process (top row) and learned reverse process (bottom row) on SAR images of sea-ice (see Section 3). The noising process starts from a real image, progressively destroying its information until only approximately pure Gaussian noise remains. The learned reverse process works in the opposite direction: starting from pure Gaussian noise, it iteratively adds small amounts of noise until a clear image is formed.

2.2. **Fast Sampling of PDMs.** The sampling time of the approximate reverse process is the main bottleneck of the original PDM formulation [SDWMG15, HJA20]. Indeed, the generation process requires going through all the reverse diffusion steps to generate new samples, each of which requires making an inference from a neural network. This is orders of magnitude slower than, *e.g.*, one pass through a trained Generative Adversarial Network (GAN). Many different sampling techniques have since been designed to simultaneously accelerate the sampling process and improve the sample quality. They notably leverage theoretical connections between PDMs and other generative models or differential equations. For instance, PDMs were shown to be equivalent to score-based generative models [SSDK+20], which allows to make use of the rich literature of score-based sampling (*e.g.*, Langevin dynamics methods) to produce efficient sampling techniques for the approximate reverse process [SE19, DVK21].
Other connections between the reverse process and ordinary differential equations (ODEs) make it possible to design fast sampling algorithms, by relying on existing, fast approximate ODE solvers [ZC22, LZB$^+$22]. Examples include DDIM [SME20] or DPM-solver [LZB$^+$22] (see Figure 5). However, such approaches rely on trade-offs between the quality of generated samples and a faster sampling time, and may sometimes generate lower-quality samples [LRLZ22].

Many other fast sampling techniques have been designed. For instance, [ND21] accelerates sampling by learning the variance of the reverse diffusion process, which allows to produce better samples with fewer steps (see Figure 6). There are also various learning-based fast sampling approaches that, for simplicity, we do not integrate into our experiments, such as progressive distillation [SH22] or truncating the reverse diffusion process [LXY$^+$22, ZHCZ22]. Such approaches can also be expensive and complex to train, and be less flexible[LZB$^+$22]. Finally, latent PDMs [RBL$^+$22], in which the forward and reverse processes are defined on a lower-dimensional latent representation of the data instead of the original high-dimensional image space, also allow for faster sampling. However, the lack of powerful deep learning feature extractors specific to the type of remote sensing data we use here (Section 3) prevents us from implementing this approach.

Here, we tested two well-established fast sampling strategies: DDIM [SME20] and ODEs [ZLCZ23], whose performance we compare to that of the traditional sampling of the original PDM formulation [HJA20]. Because these strategies have all been designed while working on natural images, it is not obvious that they would also work on non-natural images. We compare sampling strategies for the SAR dataset only (see Section 4.1), because it is the only one of the three datasets we consider which we can project to a lower-dimensional latent feature space, and for which we can thus implement quantitative performance metrics (FID and improved PR; Section 2.3).

2.3. **Evaluation Methodologies.** Evaluating the quality of network-generated images, whether in a conditional or unconditional setting, is a difficult task. Indeed, determining whether images are "realistic", in the sense that they cannot be distinguished from the set of real images, and whether they explore the full distribution of real images, is often a very subjective task for which few objective metrics exist. The difficulty lies in accurately describing the distribution of real data. Previous work on this topic has mainly focused on natural images, like CIFAR10 [HJA20], LSUN [HJA20], CelebA-HQ [SBL$^+$18, HJA20] or ImageNet [ND21].

In the natural image framework, model performance assessment relies on pre-existing models to embed images in lower-dimensional latent features spaces. The two most popular performance metrics in this respect are the Inception Score (IS) [SGZ$^+$16] and the Fréchet Inception Distance (FID) [HRU$^+$17]. Both rely on an InceptionNet model (InceptionV3) pre-trained on the ImageNet dataset [Bor22]. IS is equal to KL divergence between the conditional and marginal distributions of image labels over the generated data. FID, on the other hand, is equal to the Fréchet (*i.e.*, Wasserstein-2) distance between the multivariate Gaussian distributions fitted to the feature distribution of the real and generated images in the embedded latent space of the InceptionV3 network. Both metrics have numerous shortcomings [Bor22]. IS is especially sensitive to model parameters, does not detect intra-class mode collapse, and is biased towards the ImageNet dataset and the InceptionV3 model [BS18]. While FID is more robust to noise and small perturbations, better captures intra-class diversity and generally better correlates with human judgment [XHY$^+$18], it has high bias and relies on the assumption of gaussianity of the feature space distributions [Bor22]. The latter may not hold true, especially for non-natural images, for which the traditional InceptionV3 network may not perform well.

Another commonly used metric is the so-called Precision and Recall (PR) [SBL$^+$18]. Simply put, precision measures the degree to which generated images fall within the distribution of real images, while recall measures the degree to which real images fall within the distribution of generated images. [KKL$^+$19] introduced "improved PR" metrics that rely on explicit, non-parametric representations of the real and generated image distributions. In combination with other scores like FID, they yield important insights on the quality of generative models and their ability to cover the target (real data) distribution. Like IS and FID, PR also relies on embedding images in a lower-dimensional feature space using an image classification model. However, unlike FID, it does not assume the latent features to be Gaussian distributed, an attractive characteristic when working with non-natural images for which the literature is limited.

FID and improved PR both rely on a reliable feature extractor that, in practice, may not be available or hard to develop from scratch. Another tool that goes beyond the simple visual inspection of generated images is the semivariogram. The semivariogram measures the spatial autocorrelation of a given spatial field. Formally, given a field $Z(\mathbf{s})$, the semivariogram is defined as:

$$\gamma(\mathbf{s}_i, \mathbf{s}_j) = \mathbb{E}|Z(\mathbf{s}_i) - Z(\mathbf{s}_j)|^2$$

Assuming the field $Z(\mathbf{s})$ to be isotropic (which we do for our InSAR datasets), $\gamma$ is only function of $|\mathbf{s}_i - \mathbf{s}_j|$. For a distance $h$, the empirical semivariogram can be obtained as follows:

$$\hat{\gamma}(h) = \frac{1}{2|N(h)|} \sum_{(i,j) \in N(h)} |z(\mathbf{s}_i) - z(\mathbf{s}_j)|^2$$

where $N(h) = \{(i,j) \text{ s.t. } |\mathbf{s}_i - \mathbf{s}_j| = h \pm \delta\}$ (the set of indices such that the corresponding spatial points are separated by a distance of $h$, modulo a small $\delta$). Comparing empirical semivariograms of real and generated data therefore allows to assess whether the model accurately represents the spatial structure of the target distribution.

3. DATA AND EXPERIMENTS

We demonstrate the applicability of PDMs to generate images from three different SAR datasets that cover a wide range of applications: one dataset of SAR surface backscatter scenes, one of InSAR phase scenes, and one of InSAR-based surface deformation scenes. All datasets have a single channel, meaning that they can be thought of as grayscale data.

3.1. **SAR backscatter scenes.** We use the TenGeoP-SARwv dataset [WMT$^+$19] which consists in about 37,000 SAR backscatter scenes of varying sizes (all approximately $450 \times 500$), divided into ten geophysical categories: pure ocean waves, wind streaks, micro convective cells, rain cells, biological slicks, sea ice, icebergs, low wind areas, atmospheric fronts and oceanic fronts. All scenes were acquired by the Sentinel-1A satellite in wave mode and VV polarisation during 2016 above the world oceans. The classification was done by hand, and each scene is given a single class label.

For ease of processing, we normalise each scene from its original range to $[0, 1]$. We then divide the dataset into training (80%) and validation (20%), for each data class separately (since some of the classes have fewer data points than others). Because the size of the original data samples is large, we train a PDM on $128 \times 128$ scenes obtained by randomly cropping $256 \times 256$ scenes from the TenGeoP-SARwv samples and downscaling these to a $128 \times 128$ size. This allows us to train a reasonably-sized model on a single GPU. To show that the original resolution of the samples can be achieved in generated images, we also implement a super-resolution PDM which learns to translates low-resolution ($128 \times 128$) images to high-resolution ($256 \times 256$) ones. The principle remains the same: the input high-resolution image is progressively destroyed by adding noise, and the PDM learns to reconstruct it iteratively, but this time with conditioning on the low-resolution version of the original high-resolution image.

The TenGeoP-SARwv dataset is the only one of the three datasets we consider for which labels are available. In addition to visual inspection, we thus compute FID and improved PR for this dataset only, by fine-tuning the pre-trained InceptionV3 image classification model on our $128 \times 128$ SAR database (downscaled from $256 \times 256$ TenGeoP-SARwv cropped scenes). Note however that absolute FID values are difficult to interpret for SAR images since, unlike natural images, few if any generative models have been applied to them. Additionally, the Inception network was designed for natural images. The feature extraction step, despite the fine-tuning, may therefore not perform well enough to get reliable FID and PR scores.

3.2. **InSAR interferograms.** Second, we use an ensemble of 110,000 unwrapped InSAR interferograms cropped to a size of $128 \times 128$, obtained from SAR data acquired by the Sentinel-1A satellite over New Mexico in 2019. InSAR interferograms are obtained by subtracting SAR phase scenes between two different observations. To this end, we use the Interferometric Synthetic Aperture Radar (InSAR) Scientific Computing Environment (ISCE). SAR phase corresponds to the number of oscillation cycles that the radar wave executes as it travels from the satellite to the surface and back again. Phase is known modulo $2\pi$ only, which creates a "phase ambiguity" in the interferograms. Raw interferometric data is "wrapped", meaning that it belongs to $[-\pi, \pi]$, and thus consists of a series of interferometric fringes. This ambiguity must be resolved through an unwrapping process to access absolute phase differences.

Here, we train PDMs on the unwrapped interferograms obtained by processing with the SNAPHU (Statistical cost, Network flow Algorithm for Phase Unwrapping) algorithm. The original data range from about -100 to +100 (in multiples of $2\pi$), with most images being between -10 and 10. The samples thus wildly differ in terms of range, which can be a challenge for PDMs (see Section 5.2). Instead of training PDMs on the original data, we normalise each sample to scale between -1 and 1. Finally, since labels are not available for this dataset and we do not have access to a reliable feature extractor, we rely on visual inspection and on the comparison of (empirical) semivariograms between real and generated data to assess PDM performance.

3.3. **Noisy InSAR ground deformation scenes.** Our final dataset is an ensemble of $32 \times 32$ InSAR-based ground deformation scenes obtained from InSAR interferograms over New Mexico with the small baseline subset (SBAS) algorithm. SBAS takes in time series of unwrapped interferograms as input, and uses a time regression to estimate ground deformation over time [LXL22]. The resulting deformation field includes several kinds of noise (atmospheric noise, soil moisture noise, etc.) which partly mask the true ground deformation signal. Here again, we normalise each data sample to scale between 0 and 1.

Working with these three datasets allows to cover a wide range of SAR/InSAR data applications. In the case of SAR backscatter scenes, labelling must be performed by hand and it is thus very time-consuming to generate large datasets that can be used for massive and automated SAR image analysis. Note that the TenGeoP-SARwv dataset consists of about 37,000 cropped Sentinel-1 scenes, while the Sentinel-1 constellation alone acquires more than 60,000 scenes per month [WMT+19]. Second, InSAR interferograms are widely used *e.g.* to monitor ground deformations [*e.g.*, RLJD+21] or surface soil moisture [*e.g.*, EML20]. Specifically, unwrapping is a key step in all applications of InSAR interferometry. It consists in solving an inverse problem to reconstruct the unwrapped phase signal from the observation of the phase modulo $2\pi$. It is traditionally performed by path-following or optimization-based algorithms such as SNAPHU. However, deep learning approaches have recently shown promise [YCF+19, WLK+19, QWW+20, YYS+20, SGG20, WWW+21, PDS21, ZCP+21, SRG22, MRP22, HMW+22, ZL22] and potentially offer a scalable and computationally efficient solution for global InSAR Earth monitoring. One of their main limitations is the size of training databases, currently limited by the need to unwrap interferograms with algorithms such as SNAPHU. Converting an unwrapped signal to a wrapped signal is instantaneous (one just has to take the modulo $2\pi$ phase residual). Generating large numbers of unwrapped interferograms (and associated wrapped scenes) could therefore help augment the performance of deep learning unwrapping algorithms by building much larger databases than are currently used. Finally, noisy time series of InSAR-based ground deformation are used as input to denoising algorithms that extract fine-scale ground deformation maps. Generating large amounts of such series is key to improving the robustness of deep learning-based denoisers [*e.g.*, RLJD+21, SWEAM22a, OGC+22, SWEAM22b, SSC22]. Denoising is a crucial step of InSAR-based deformation analysis. It is particularly challenging due to the many sources of significant noise to be found in interferograms, like the atmosphere and changes in surface scattering properties due to soil moisture or vegetation [AS15].

4. RESULTS

4.1. **SAR backscatter scenes.** We begin with our experiment using the SAR backscatter TenGeoP-SARwv dataset [WMT⁺19]. We find that a PDM trained with 2000 diffusion time steps and which learns the variance yields visually satisfactory results after about 50 epochs ($\approx$ 1.6 million examples), and the validation loss stagnates after around 70 epochs ($\approx$ 2.2 million examples). Examples of class-conditional generated 128 × 128 images with this model are shown on Figure 3 for eight of the 10 classes. The generated samples appear visually indistinguishable from the original data, and exhibit the characteristics of their corresponding class (*e.g.* localised icebergs for class L or isolated rain cells for class G).

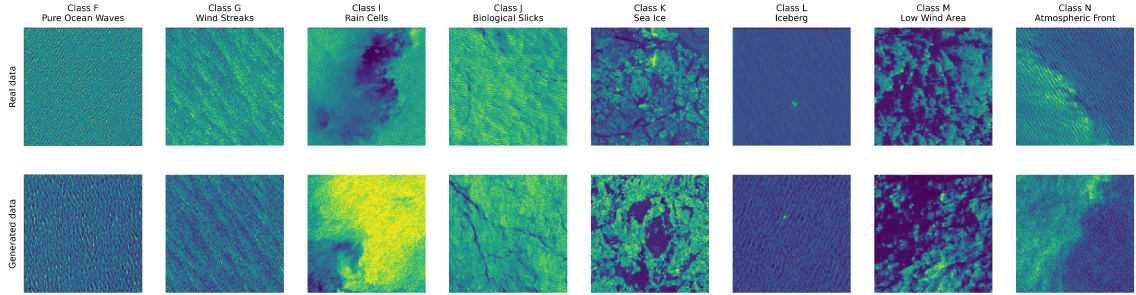

FIGURE 3. Examples of SAR images for 8 of the 10 TenGeoP-SARwv classes: top row, examples from the TenGeoP-SARwv database, and bottom row: generated with our PDM model.

The subsequent image enhancement by the super-resolution model leads to high-quality images that are hard to distinguish from the original data (Figure 4). Some images, however, may exhibit excessive fine-scale granularity (*e.g.*, bottom left example on Figure 4) that could result from an improper UNet architecture (see Section 5.1).

We compute FID and improved PR for various numbers of diffusion time steps to further compare real and generated data, and to assess the performance of our models and the various sampling strategies. The fine-tuned InceptionV3 classification model trained on this dataset to extract the latent representation of images achieves a relatively good performance, with an accuracy of about 94%. We could not achieve higher values, likely due to (i) the small size of the TenGeoP-SARwv database (37,000) compared to that used to train the InceptionV3 model (ImageNet with $\approx$ 1.3 million examples), and (ii) the relative loss of class distinctiveness due to randomly cropping scenes to a 256 × 256 size. We nevertheless use this fine-tuned model and crop it at the level of its final average pooling features to extract a latent space of dimension 2048. Figures 5 and 6 show PDM performance – in terms of FID (Figure 5) and Precision and Recall (Figure 6) – as a function of the number of sampling diffusion time steps (which correlates to the time required for sampling). We compare two PDMs: one which learns the variance ("variance"; results shown in Figure 3), and one which doesn't ("original"). Both are trained with 2000 diffusion time steps and for the same number of epochs (70). For the "variance" model, we also compare traditional sampling with DDIM ("DDIM_var") sampling.

First, we see that increasing the number of diffusion time steps enhances the quality of generated samples for the original sampling strategy. We find sharp decreases in FID and increases in precision and recall as more diffusion time steps are added. The absolute improvement in FID by moving from 10 to 500 steps is largest for the "original_novar" configuration (from a FID of 245 to 25). This is not surprising: the closer the number of diffusion steps is to that on which the model was trained, the better the images should be. Generated images with 10-50 diffusion steps are clearly still too noisy and more time steps are required to generate fine-scale structures (see right-hand panel of Figure 5). DDIM sampling, however, shows very poor performance, regardless of the number of diffusion steps. FID and recall remain about constant (FID $\approx$ 120 and recall $\approx$ 0.35) for all tested diffusion time steps, and only precision increases from 0.07 to 0.25. We see on the right-hand panel of Figure 5 that DDIM sampling tends to generate images that are too noisy or

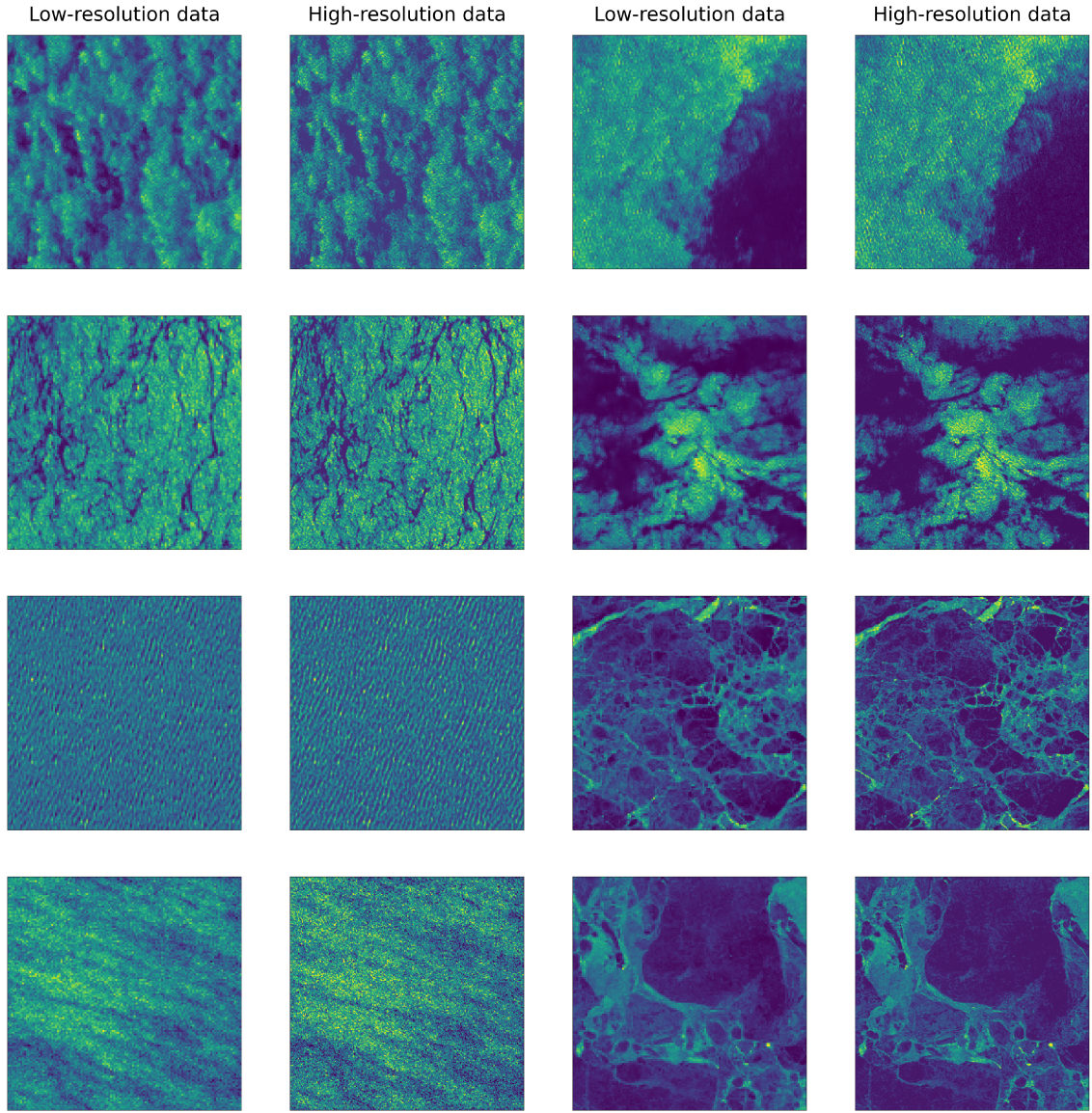

FIGURE 4. Examples of results for the super-resolution model: low-resolution ($128 \times 128$) images generated by a PDM (first and third columns) and their high-resolution ($256 \times 256$) version obtained by a super-resolution PDM (second and fourth columns).

too smooth, and fails to reach the level of structural detail attained by original sampling. We find similar results when applying the DPM-solver sampling approach (not shown), despite its good results on the MNIST dataset (Figures 10 and 11)

Second, for a given number of time steps, the model which learns the variance performs (almost always) better than the model which doesn't. In other words, the learned-variance model requires fewer diffusion steps at sampling to generate images of a given quality. For instance, 50 diffusion time steps are sufficient to yield samples of equal quality as the model which doesn't learn the variance with 100 diffusion steps (which takes about 1.5 times longer). This result is consistent with Nichol and Dhariwal (2021) [ND21] who introduced the learned-variance models and found that these could generate high-quality samples with small numbers of diffusion steps on natural images. Beyond about 200 diffusion time steps, however, both models perform about equally well with the original sampling method. FID converges to a value of ≈25-30 and precision and recall to ≈0.5-0.6. With 500 diffusion steps, the learned-variance model has lower precision but slightly better recall, and overall its performance seems comparable to that of the no-variance model. Note however that since fewer inferences are required during sampling in the no-variance model, the latter requires less time (about 25% less) to generate a given number of images as the learned-variance model.

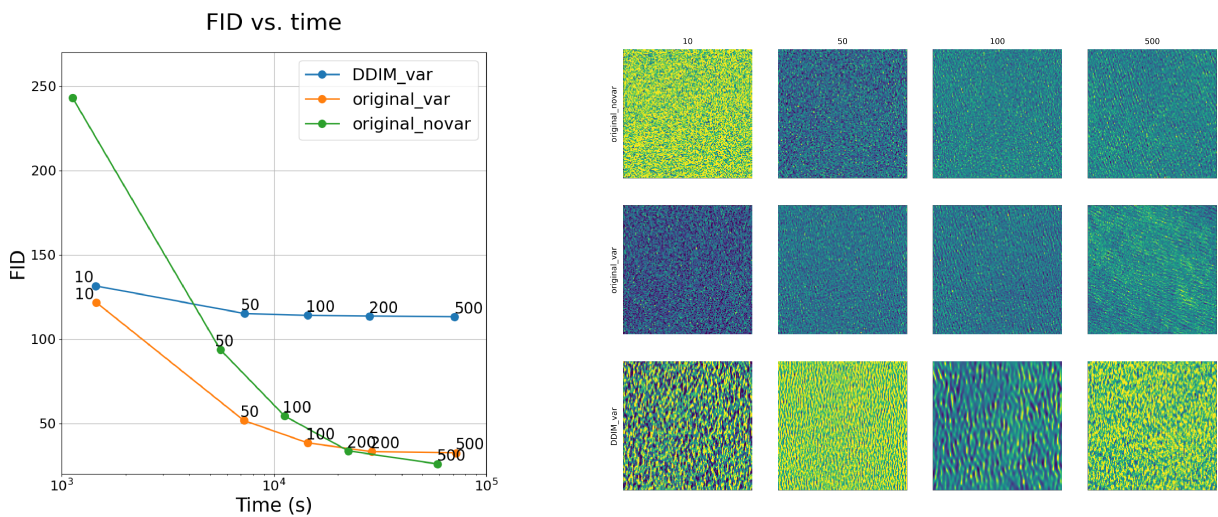

FIGURE 5. Left panel: Quality of generated images as a function of sampling time, sampled from PDMs trained on the TenGeoP-SARwv dataset, with ("original_var") and without ("original_novar") learning the variance, and with original and DDIM-based sampling, measured in terms of FID. The number of sampling diffusion time steps (10, 50, 100, 200 and 500) is indicated next to each data point. Each point corresponds to 10000 generated images, and therefore to approximately the same sampling time for each of the three curves. Right panels: examples of generated images (pure ocean waves) with different numbers of diffusion time steps (10, 50, 100 and 200).

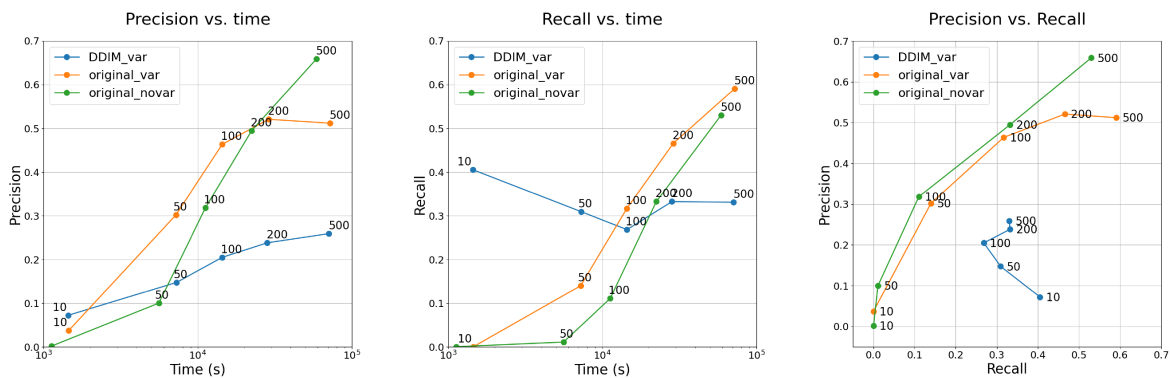

FIGURE 6. Same as the left panel of Figure 5, but for the improved PR metrics: precision vs. sampling time (left panel), recall vs. sampling time (middle panel) and precision vs. recall (right panel).

4.2. **InSAR interferograms.** We now turn to the results obtained on the InSAR unwrapped interferogram dataset. Figure 7 shows examples of real images from the test dataset and of generated images with our trained PDM. The model was trained with 1000 diffusion steps over 80 epochs ($\approx$ 3.4 million data samples).

Generated samples are visually very similar to the real ones, even displaying topography-like features also present in the real data (see for instance the third and fourth bottom panels in Figure 7). While training samples were all scales to $[-1, 1]$, we find that the PDM struggles to produce images with that exact same range. By a posteriori rescaling generated images to $[-1, 1]$, we find that the semivariograms of the real and generated data are highly similar (Figure 8-a), suggesting that the PDM succeeds in capturing the spatial structure of the data. The nugget (the intercept of the semivariogram at a lag distance of almost zero) is zero for both sets, and while the size of the samples ($128 \times 128$) is insufficient to reach a distance of total decorrelation, the slope of the semivariogram and its maximum value are also comparable. We also show in appendix (Figure 12) the semivariogram calculated with the generated data before rescaling. It corroborates that the generated data has too little variance compared to the test data, hence smaller semivariogram values across the range of distances.

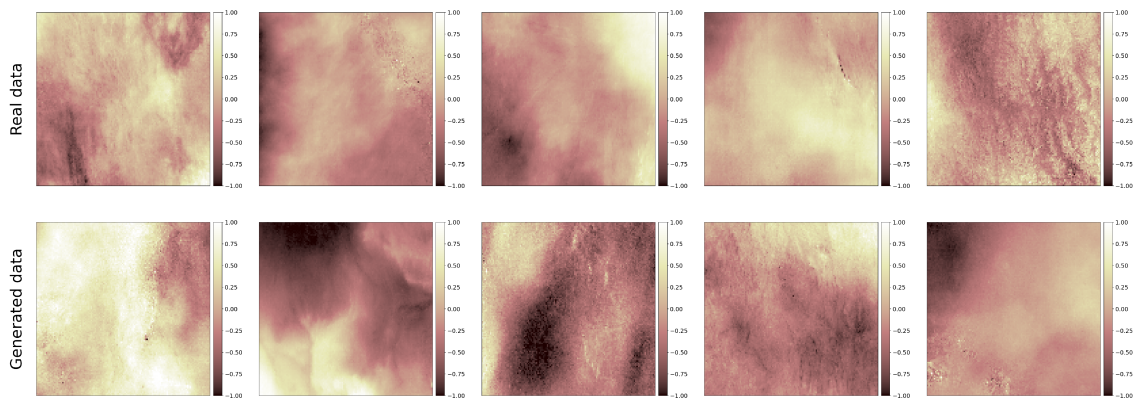

FIGURE 7. Examples of unwrapped InSAR interferograms from the validation dataset (top row) and generated using 500 diffusion time steps by a PDM trained with 1000 diffusion time steps (bottom row).

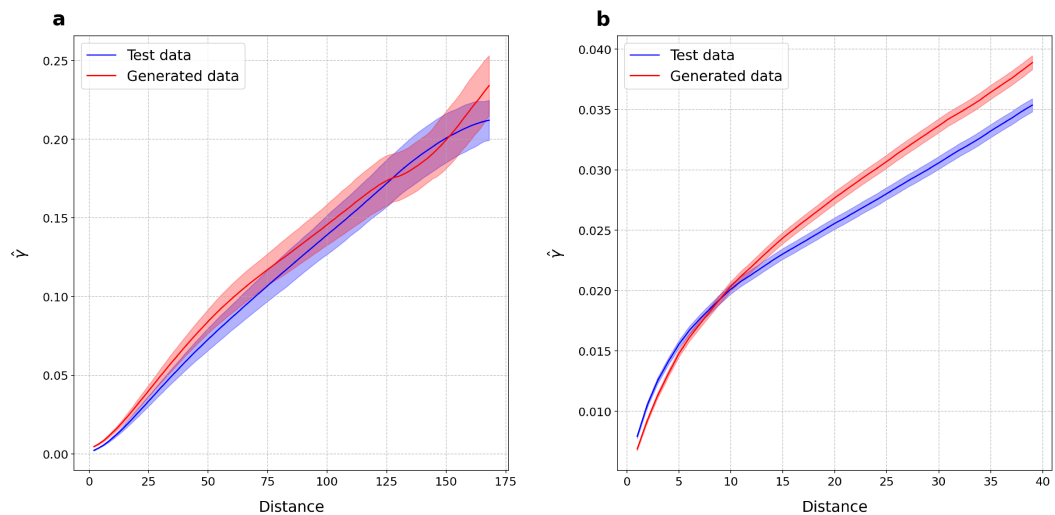

FIGURE 8. Comparison of semivariograms between test and generated data for (a) the unwrapped InSAR interferograms (obtained with 250 data samples of size $128 \times 128$) and (b) the InSAR ground deformation data (obtained with 10000 data samples of size $32 \times 32$). In both panels the 95% confidence interval, obtained from the range of semivariogram values across the samples, is indicated by the colored shading around each curve.

4.3. **Noisy InSAR ground deformation scenes.** Last, we discuss the results obtained on the InSAR noisy ground deformation dataset. Figure 9 shows a random selection of real and generated scenes, using a PDM trained with 1000 diffusion time steps and which learns the data variance. Again, by eye, generated and real samples appear very similar. To go further and obtain a more quantitative comparison, we show on Figure 8-b the two semivariograms (real and generated data). The two curves are similar; in particular, the nugget is almost equal. Here again, the image size ($32 \times 32$) is too small given the spatial correlation of the data for the semivariogram to flatten out and to compare the sill and range of the two curves. However, it appears that the generated data tend to have too little spatial covariance (*i.e.*, too high semivariogram values) compared to the real data beyond a distance of $\approx 15$. The difference is small but significant. This problem is not related to the PDM failing to generate data with the right $[0, 1]$ range – on the contrary, in this example, the PDM generates samples with the correct range (Figure 9), and tends to underestimate the covariance at large distance.

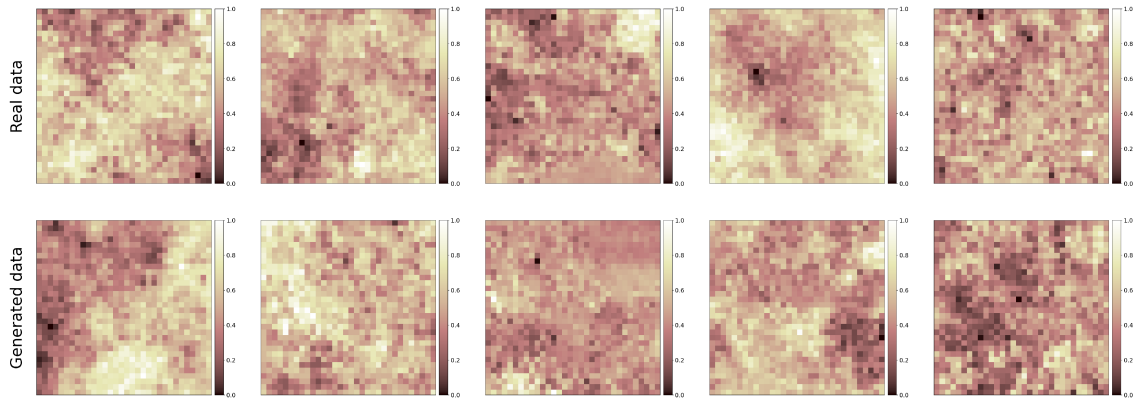

FIGURE 9. Examples of InSAR-based noisy ground deformation snapshots from the training database (top row) and generated using 500 diffusion time steps by a PDM trained with 1000 diffusion time steps (bottom row). Here, the real deformation maps have been normalized to $[0, 1]$, but the PDM can seamlessly accommodate different ranges of data, see Section 5.

## 5. Discussion and Conclusion

5.1. **Model convergence and hyperparameter selection.** PDM convergence during training is very much function of the size and complexity of the modeled data. In all cases, they frequently require millions of iterations to converge [DN21]. This is what we found *e.g.* on the $128 \times 128$ SAR dataset for which visually satisfactory results appeared after around 50 epochs, or 1.6 million data samples. As usual in such cases, model convergence can be assessed by calculating the corresponding loss on a validation data set. Loss calculation can be very time-consuming, however, and one might prefer to evaluate model performance a posteriori rather than during training. Note though that PDMs succeed in converging satisfactorily even on rather small training datasets. We trained a PDM on a subset of the TenGeoP-SARwv dataset (single class K, sea-ice, with about 4000 single scenes) and found that the model achieved a similar performance to the one trained on the full dataset (with about 10 times as many data samples).
Like Nichol and Dhariwal [ND21], we found that model performance tended to increase with model complexity and training time, especially for datasets with complex structure/texture like TenGeoP-SARwv. PDMs can quickly reach into the millions of parameters, however, and saturate single GPUs. Two-step models (one generative model coupled to a super-resolution model) can be a good compromise to generate high-quality, high-dimension images, instead of a single generative model that would require more RAM to train. Model complexity is expressed, among others, by the number of diffusion time steps, and the UNet model which learns the noise structure. For simple datasets like MNIST (small image size and low image complexity), a few dozen of diffusion steps are sufficient in practice, but for radar-based imagery, we found that several thousands of time steps were required to yield satisfactory results. We show above the results for PDMs trained with 2000 diffusion steps, but going up to 4000 slightly improves the results (not shown). As to the UNet model, its architecture seems critical to the performance of the resulting PDM. The UNet must be able to capture the structure and texture of the target data. Traditional UNet architectures work well with natural images like ImageNet, but we see in our SAR results that fine-scale details in non-smooth classes (like sea-ice, see Figure 3) remain sharper in the real data. Similarly, the PDM trained on the noisy ground deformation scenes struggled to generate data with sufficient spatial covariance (Figure 8-b). Both cases indicate that generated images may retain too much unstructured noise. Further work may thus be required to design UNet architectures better suited to radar-based imagery and improve PDM generative performance.

On a final note, many deep learning tasks of InSAR data processing (*e.g.*, unwrapping of interferogram time series [HZ07]) or InSAR data interpretation [RLJD+21] take time series as input. The PDM sampling framework can be easily adapted by converting time series into multi-channel images or by parametrizing the Markov transitions of the diffusion process with recurrent neural networks [YSM22] instead of UNets, as is currently the case.

5.2. **Influence of data normalisation and range.** Data normalisation and range appear very important to the convergence and performance of PDMs. There are two aspects of the data to consider in this regard: the diversity of ranges of individual data samples, and the average magnitude of the data samples.

The first aspect (range of individual data samples) is challenging because PDMs tend to produce samples with similar ranges of values (*e.g.*, between -1 and 1). Consequently, they have trouble to accurately sample from the target distribution when individual data samples can have very different ranges (*e.g.*, most samples ranging from -1 to 1 but a few between -10 and 10). This occurs frequently in remote-sensing datasets – examples include RGB scenes with varying brightness depending on cloud cover or time of acquisition, or InSAR interferograms with varying atmospheric-induced noise (see Section 3). By contrast, in the literature, PDMs have mostly been applied to natural images which range between 0 and 255 (RGB color scale). In fact, most images in common training datasets like ImageNet [ND21] explore most or all of the $[0, 255]$ range, and very few are restricted to only one side of the range (*i.e.*, very dark or very bright images). This creates a homogeneity that seems to help PDMs to converge and makes sampling easier.

We found that when trained on unnormalized data with large differences in individual ranges, PDMs tend to generate samples at the extremes of the range distribution. This led to too much variance in generated samples compared to the real ones, even if the structure of generated samples may be highly realistic. It may also happen that, when trained on normalised data, PDMs fail to generate normalised images – for instance, in the unwrapped interferogram case (section 4.2), while all training images were scaled to $[-1, 1]$, generated samples often explored a smaller range (*e.g.*, $[-0.5, 0.5]$).

During sampling, the noise amplitude can be controlled to force the generated sample to remain within prescribed bounds. However, in the case of widly varying data ranges, it isn't obvious how to control properly the amplitude of the noise during sampling to force generated images to better fit to the distribution of data ranges. This is why we trained all our PDMs on normalized datasets (values in each image being restricted to $[-1, 1]$ or $[0, 1]$). Generated images can then be later rescaled to the target unit by sampling from the distribution of normalising coefficients in the training dataset. This admittedly remains an imperfect solution (the normalising coefficients may not be independent from the complexity of the image) for which further improvements in PDM sampling are likely required.

The second aspect (average data magnitude) is also important to consider. For a given number of diffusion time steps, the larger the data magnitude is, the higher the noise variance to be added at each time step must be. This creates difficulties in practice. First, for PDMs where the variance is not learned (the noise variance being set a priori with a noise scheduler), this can lead to bad model convergence. Second, for PDMs which do learn the variance, model convergence is much longer. Figure 13 illustrates these difficulties on the $32 \times 32$ MNIST dataset, where we trained PDMs on the original dataset (scaled to $[-1, 1]$) and on a version rescaled to $[-100, 100]$. The model trained on the original data performs much better and takes considerably less time to converge to satisfactory results. The learned-variance model on the rescaled dataset does yield somewhat recognizable results, but of much lesser quality for the same training time. As to the model which doesn't learn the variance, results are extremely poor.

5.3. **Assessing model performance.** Assessing the performance of image generation models is a difficult task for which few objective, quantitative metrics exist. Existing approaches, some of which we implemented in this paper, typically rely on a feature extraction step to move from the image space to a latent space of lower dimensionality. This extraction is usually performed with classification models like InceptionV3 or VGG16. This requires working with labeled datasets, which is not always the case. Additionally, most (if not all) commonly-used classification models were developed for natural images and trained on extensive datasets like ImageNet ($> 1$ million examples). There is consequently no guarantee that fine-tuning such models will work on non-natural, single-channel images for which available datasets are potentially very small (such as TenGeoP-SARwv).

We saw in our example of SAR backscatter scenes that despite the fine-tuning, the classification model only reaches 85% accuracy. Additionally, the poor precision and recall values, despite the good visual quality of generated images, suggests that the feature extraction is unreliable. The FID and PR values shown here (Figures 5 and 6) should therefore be interpreted in relative rather than absolute terms, in the same way as statistical skill scores like the Akaike information criterion.

When working with non-natural images (whether labeled or not), one could think instead of extracting features with VAEs. VAEs are models tasked with reconstructing images from a smaller-dimensional set of latent features (and can incidentally be used in generation mode, like PDMs). We have not done this in the present work, but it may be a promising approach to apply traditional performance scores like FID or PR to non-natural images with more confidence.

**Conclusion**

In this paper, we explore the ability of PDMs to generate realistic images from various datasets of radar-based remote sensing imagery, which cover a wide range of practical applications. We also provide a versatile and robust (in terms of training parameters) open-source code to further explore how PDMs can be applied to any such dataset of non-natural images.

The main advantage of PDMs is their versatility and ease of training. Little engineering is required to have them capture complex image distributions, even in remote sensing. In addition, they are less prone to mode collapse than their competitors, especially GANs. The sampling time remains a significant bottleneck of PDMs. Generating new images indeed takes much longer than for other generative models. We find that efficient sampling approaches proposed for natural images, like DDIM or DPM-solver, fail to produce quality images on the SAR dataset. PDMs therefore remain very slow in generation mode to be used in real-time applications, though they offer an attractive alternative to GANs for offline image generation (*e.g.*, labeled database creation).

Classic performance metrics (FID and improved PR) nevertheless remain unsatisfactory to accurately assess PDM performance in non-natural image generation. Available feature extractors were indeed pre-trained on natural image datasets, and appear insufficient to correctly characterise radar-based images. Designing feature extractors tailored to remote-sensing imagery is therefore key to the development of PDM in the field. Such extractors will also make it possible to experiment latent PDMs [RBL+22], which learn the reverse diffusion process on a latent (lower-dimensional) representation of images, and considerably reduce the sampling computational bottleneck.

The automated analysis of remotely-sensed imagery, notably radar-based data, remains challenging due to the small size of available annotated datasets to train learning algorithms. Automated object detection (*e.g.*, ships), SAR scene classification, or InSAR phase unwrapping algorithms, indeed require extensive hand-labeled datasets that are very time-consuming to obtain, and may not necessarily allow fitted models to work well with different satellites. Current databases remain relatively small, of the order of a few ten of thousand examples [*e.g.*, YCF+19, MRP22]. This is why fast generative methods – often based on GANs – have been developed to circumvent this limitation [*e.g.*, RRHP19, ZZW$^+$20, LXS$^+$22, LQJ$^+$22, WSL$^+$23]. PDMs were recently introduced to the field of generative models and have already shown promising, if not state-of-the-art results on natural images. Our results show that PDMs also hold much promise on complex radar-based datasets, and we expect them to improve many learning challenges in remote sensing beyond the three main use cases presented in this paper.

**Acknowledgments.** This work was funded by NASA under the NSPIRES grant 80NSSC22K1282. B.R.-L. also acknowledges the JSPS LEADER funding.

**Author contributions.** A.T. and T.K. conceived the study, wrote the code, processed all results and wrote the manuscript. C.H. and B.R.-L. processed the InSAR data and B.R.-L. secured the funding for this work.

5.4. **Code and data availability.** Our code is available at https://github.com/thomaskerdreux/PDM_SAR_InSAR_generation under the MIT license. The TenGeoP-SARwv dataset [WMT$^+$19] is provided by IFREMER and publicly available at https://doi.org/10.17882/56796.

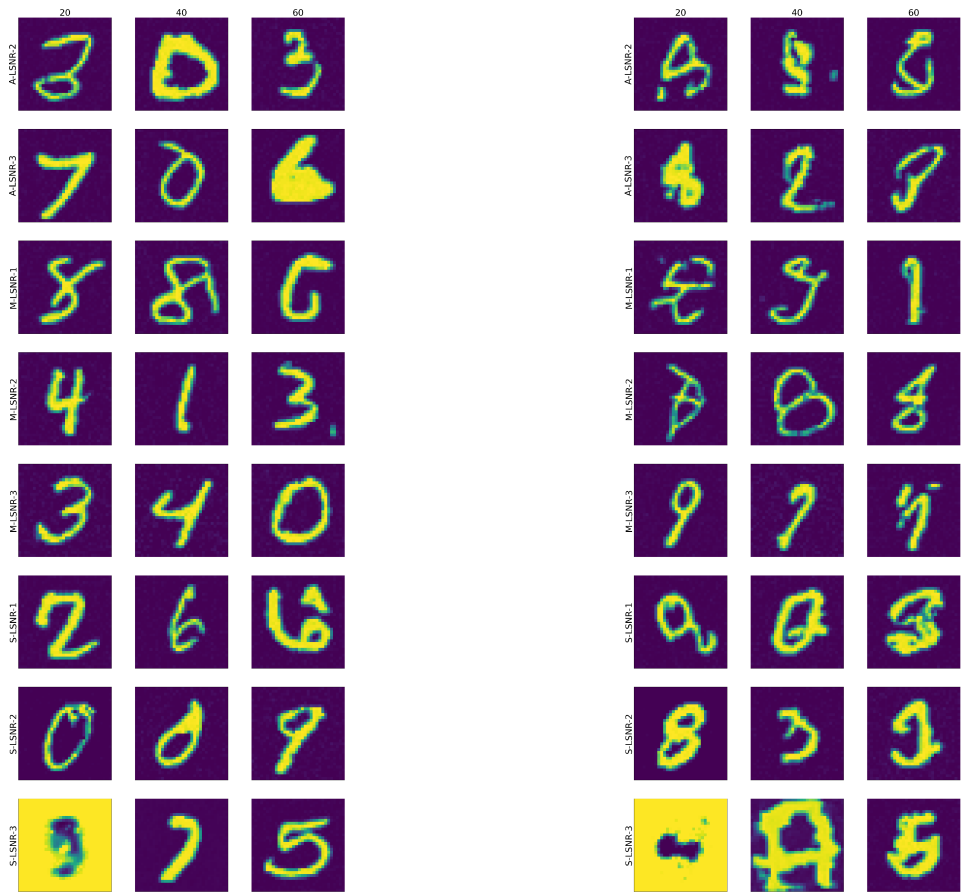

FIGURE 10. Comparison of randomly sampled images generated according to various DPM sampling parameters using PDMs trained on MNIST. The left panels show results for an unconditional PDM model, and the right panels show results for a conditional PDM model. 'S' stands for single-step, 'M' for multistep, and 'A' for adaptive, and 'LSNR' stands for 'logSNR' (see [LZB+22] for details). Note that Lu et al. (2022) [LZB+22] recommend using the 'logSNR' parameters for low-resolution datasets such as MNIST. In this example, we observe that some combinations of parameters lead to unstable results. For instance, the single-step method with order 1 for the unconditional model leads to poor results. Nevertheless, DPM sampling is a fast generative solution for the PDM models trained on MNIST. In the case of the SAR dataset, however, we observe frequent instabilities that corrupt the quality of sampled images.

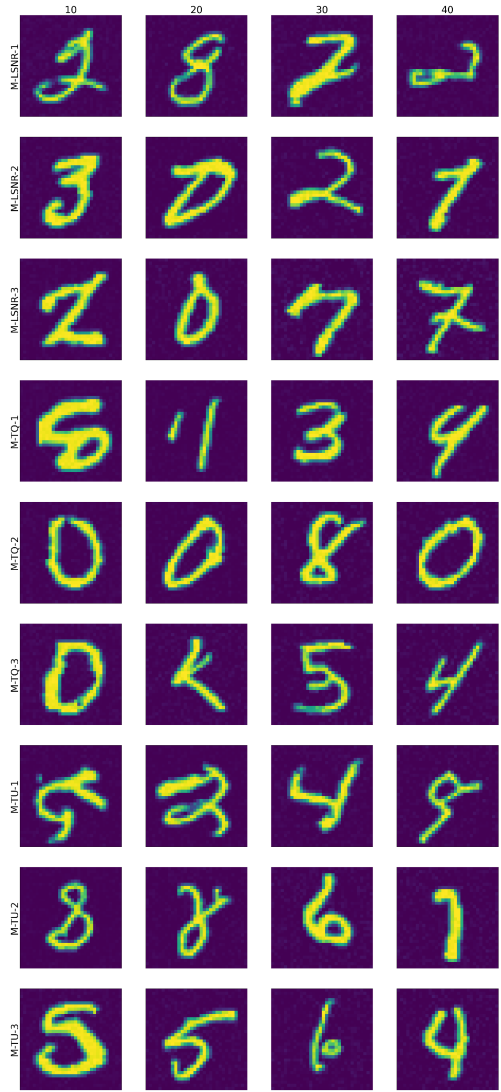

FIGURE 11. Here, we plot samples from the DPM solver with different *skiptime* parameters for an unconditional PDM model for MNIST, using only the multistep method. Note that Lu et al. (2022) [LZB+22] recommend using the 'logSNR' as a *skiptime* parameter for low-resolution datasets such as MNIST. 'TU' stands for 'time uniform'; 'LSNR' for 'logSNR', and 'TQ' for time quadratic. We observe that this parameter does not have much effect on the sampling quality.

Geolabe LLC, Los Alamos, NM, USA
*Email address*: atuel@geolabe.com

Geolabe LLC, Los Alamos, NM, USA
*Email address*: thomask@geolabe.com

Geolabe LLC, Los Alamos, NM, USA
*Email address*: claudiah@geolabe.com

Geolabe LLC, Los Alamos, NM, USA
*Email address*: bertrandrl@geolabe.com

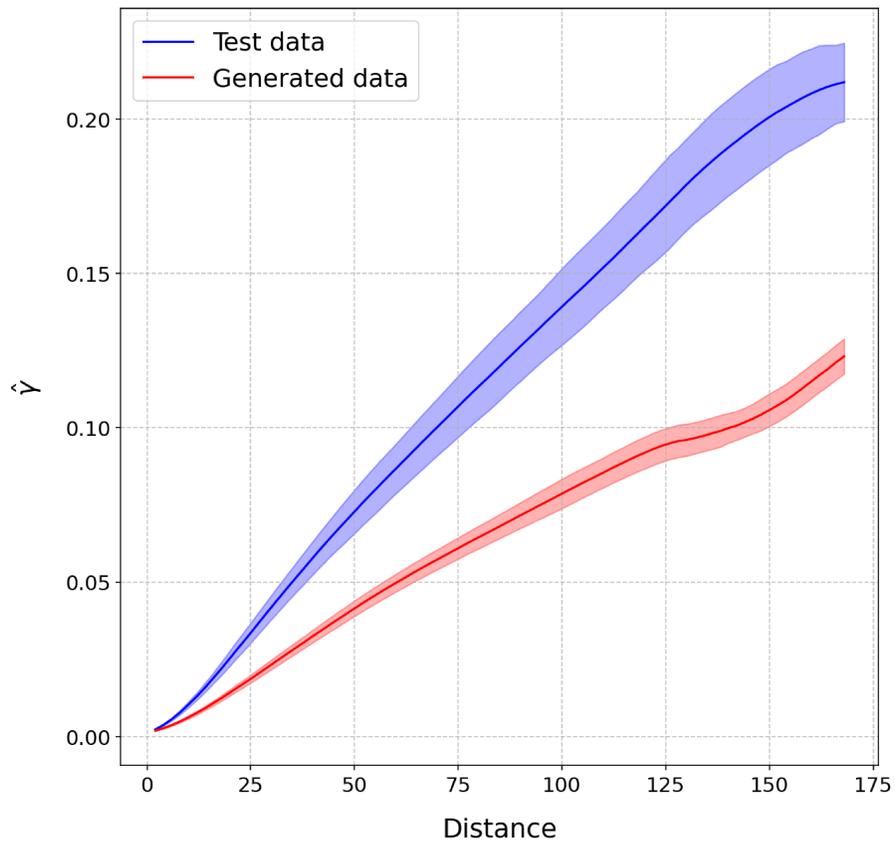

FIGURE 12. Comparison of semivariograms between test and unscaled generated data for the unwrapped InSAR interferograms (obtained with 250 data samples of size $128 \times 128$). See Figure 8-a for comparison with the scaled generated data. The 95% confidence interval, obtained from the range of semivariogram values across the samples, is indicated by the colored shading around each curve.

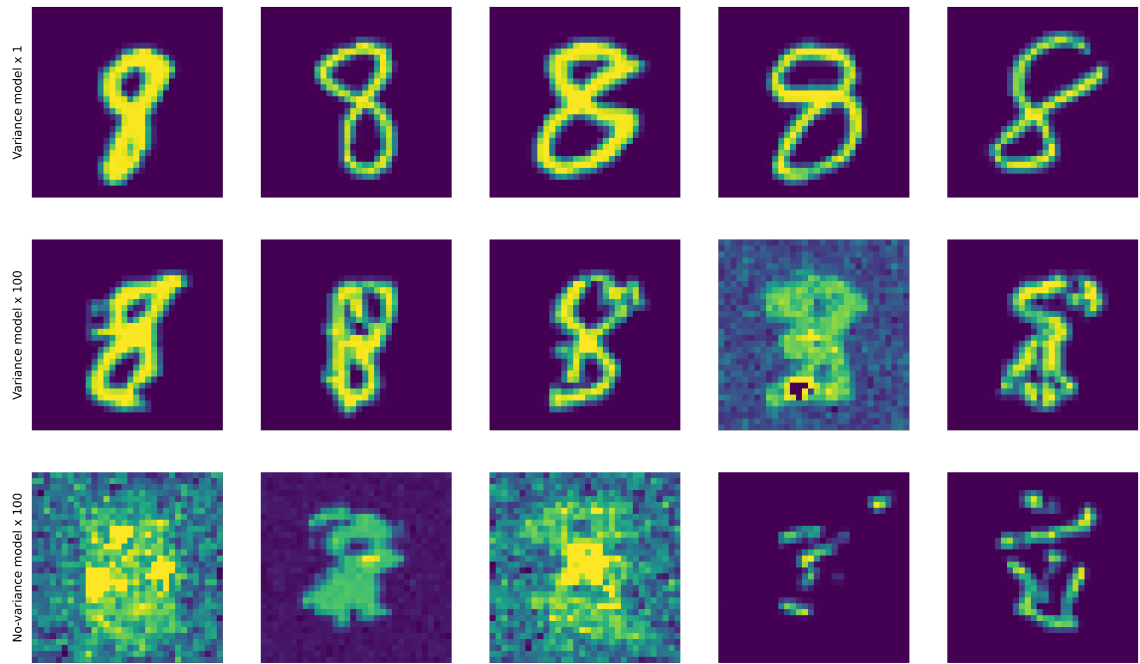

FIGURE 13. Illustrating the influence of data normalisation on PDM convergence and performance. Generated samples (class "8") from PDMs trained on the MNIST dataset: original dataset (top row) and dataset scaled by a factor of 100 (middle and bottom rows). Noise variance is learned by the PDM in the top and middle rows. All models were trained with 100 diffusion time steps for 30 epochs ($\approx 250{,}000$ data samples).